\documentclass[conference]{IEEEtran}
\usepackage[latin9]{inputenc}
\usepackage{array}
\usepackage{float}
\usepackage{multirow}
\usepackage{amsmath}
\usepackage{amssymb}
\usepackage{graphicx}
\PassOptionsToPackage{normalem}{ulem}
\usepackage{ulem}

\makeatletter

\providecommand{\tabularnewline}{\\}
\floatstyle{ruled}
\newfloat{algorithm}{tbp}{loa}
\providecommand{\algorithmname}{Algorithm}
\floatname{algorithm}{\protect\algorithmname}

\IEEEoverridecommandlockouts
\usepackage{cite}
\usepackage{amsfonts}\usepackage{algorithmic}
\usepackage{textcomp}
\usepackage{xcolor}
\def\BibTeX{{\rm B\kern-.05em{\sc i\kern-.025em b}\kern-.08em
    T\kern-.1667em\lower.7ex\hbox{E}\kern-.125emX}}

\makeatother

\begin{document}
\title{PYPM-GGD: Pitman-Yor Process Mixture with Generalized Gaussian Density
using ADAM}
\author{\IEEEauthorblockN{Kart-Leong Lim} \IEEEauthorblockA{\textit{Institute of Microelectronics, A{*}Star} \\
 Singapore \\
 lkartlx@gmail.com}}
\maketitle
\begin{abstract}
Large scale Bayesian nonparametrics (BNP) learner such as Stochastic
Variational Inference (SVI) can handle datasets with large class number
and large training size at fractional cost. Like its predecessor,
SVI rely on the assumption of conjugate variational posterior to approximate
the true posterior. A more challenging problem is to consider large
scale learning on non-conjugate posterior. Recent works in this direction
are mostly associated with using Monte Carlo methods for approximating
the learner. However, these works are usually demonstrated on non-BNP
related task and less complex models such as logistic regression,
due to higher computational complexity. In order to overcome the issue
faced by SVI, we develop a novel approach based on the recently proposed
constant stepsize stochastic gradient ascent to allow large scale
learning on non-conjugate posterior. Unlike SVI, our new learner does
not require closed-form expression for the variational posterior expectatations.
Our only requirement is that the variational posterior is differentiable.
In order to ensure convergence in stochastic settings, SVI rely on
decaying step-sizes to slow its learning. Inspired by SVI and Adam,
we propose the novel use of adaptive stepsizes in our method to significantly
improve its learning. We show that our proposed methods is compatible
with ResNet features when applied to large class number datasets such
as MIT67 and SUN397. Finally, we compare our proposed learner with
several recent works such as deep clustering algorithms and showed
we were able to produce on-par or outperform the state-of-the-art
methods in terms of clustering measures. 
\end{abstract}


\section{Introduction}

Bayesian nonparametrics (BNP) is widely used in image processing,
video processing and natural language processing. A common task in
BNP also known as model selection is to automatically estimate the
number of classes to represent an unlabelled dataset while clustering
samples (or label) accordingly. A widely used BNP is the Variational
Bayes Dirichlet process mixture \cite{blei2017variational,Hoffman2013}.

In the past, approximate learning for BNPs is mainly based on Variational
Inference (VI) where it iteratively repeats its computational task
(or algorithm) on the entire dataset, also known as batch learning
\cite{blei2003latent,bishop2006pattern}. Today, most large scale
BNP learners such as Stochastic Variational Inference (SVI) \cite{Hoffman2013,Paisley2015,kucukelbir2017automatic}.
The latter repeats its computational task on a smaller set of randomly
drawn samples (or minibatch) each iteration. This allows the algorithm
to ``see'' the entire dataset especially large datasets when sufficient
iterations has passed. However, both SVI and VI rely on closed-form
solution to work. Thus, they are limited to to conjugate posteriors.
To remove this constraint, several recent works turn to Monte Carlo
gradient estimator (MC) to approximate the expectation (or gradient)
of non-conjugate posterior. However, MC algorithms come at an expensive
cost since it require generating samples from the approximated posteriors.
Moreover, such works are usually confined to binary classifier such
as logistic regression \cite{welling2011bayesian,mandt2017stochastic,paisley2012variational,ranganath2014black}
or Gaussian assumptions \cite{rezende2015variational,kingma2014stochastic}
and mainly demonstrated on datasets with smaller class numbers such
as MNIST or UCI repository. Thus, the MC approach described above
are more suitable to relatively simpler parameter inference problems.

Due to the recent paradigm shift towards deep ConvNet (CNN) \cite{krizhevsky2012imagenet,he2016deep}
and generative networks \cite{tian2017deepcluster,xie2016unsupervised,li2018discriminatively,song2013auto,kingma2014stochastic,goodfellow2014generative},
it is very rare to find newer works following the pipeline of SVI
or MC since CNN and generative networks do not specifically deal with
model selection or unsupervised class prediction. 

The main problems faced by SVI and MC are: 
\begin{verse}
1) SVI - The approximate posterior must come from the conjugate exponential
family e.g. Gaussian-Gamma.

2) MC - Not scalable since method requires generating samples from
the approximated posteriors which is expensive.
\end{verse}
The contributions in this work are: 
\begin{verse}
1) VI without closed form - We use stochastic gradient ascent instead
of closed form coordinate ascent for VI as similarly in \cite{mandt2017stochastic}. 

2) Adaptive stepsize - Inspired by Adam, we use decaying stepsize
on both 1st and 2nd order moment of gradient for optimizating stochastic
gradient ascent.

3) Non-conjugate posterior - We introduce the generalized Gaussian
density as our mixture model. There is no closed form solution for
the VI of this model. 
\end{verse}
We test the performance of our proposed learner on large class number
datasets such as MIT67 and SUN397. Due to using deep ConvNet features
(ResNet18), we also reported better results than most recent literature
baselines. 

This paper is organized as follows: Firstly, we recall VMM \cite{lim2018fast}
for conjugate posteriors and discuss why it cannot work on nonconjugate
posterior. Next, we propose using SGA for learning non-conjugate posterior.
We further improve this learning with an Adam like stochastic optimization.
We then present an algorithm that iteratively learns all the hidden
variables of PYPM in a typical VI fashion. Lastly, we perform a study
on several datasets including the more challenging MIT67 and SUN397
to evaluate the performance of our proposed method and enhancements.
Finally, we include comparison with latest published works citing
the datasets we use.

\subsection{Related Works}

SVI do not involve actual computation of SGA. Instead, SVI parameters
are initially computed by closed-form solution \cite{Blei2006} and
then corrected via a weights biasing step. The weights follow a decay
that gradually bias towards earlier computed values. SVI is recently
demonstrated on BNP models with conjugate posterior such as the hierarchical
Dirichlet Process topic model \cite{Hoffman2013} and on large datasets
as large as 3.8M samples and 300 classes. 

On the other hand, MC methods use SGA to perform learning. Thus MC
methods works on non-conjugate posterior. SGA was also recently discussed
in \cite{mandt2017stochastic} for learning approximate posterior.
However, the authors mainly use SGA with constant stepsize for learning.
Some notable works in this area include the black box VI \cite{ranganath2014black},
VI with stochastic search \cite{paisley2012variational}, the stochastic
gradient variational Bayes \cite{kingma2014stochastic} and the stochastic
gradient Langevin dynamics \cite{welling2011bayesian}. 

\section{Problem Statement }

We present the problem of learning a non-conjugate posterior for model
selection in Bayesian nonparametrics. We introduce a variant of Gaussian
mixture model (GMM) that exist outside the exponential family distribution.
Our model of choice is the Pitman-Yor process mixture (PYPM) with
a generalized Gaussian mixture model (GGD). The GGD is a versatile
3 parameters model with mean, shape and scale parameters $\left\{ B,s,\rho\right\} $.
It can model the non-Gaussianity assumption for datasets. For simplicity,
we only focus on the following assumption for PYPM, which has the
simplest form for non-conjugate posterior i.e. by treating $\left\{ s,\rho\right\} $
as constant variables for GGD

\begin{equation}
\begin{array}{c}
x\mid B,z\sim\mathcal{GGD}\left(B_{k}\right){}^{z_{nk}}\\
B\sim\mathcal{N}\left(m_{0},\lambda_{0}\right)\\
z_{nk}\mid v_{k}\sim Mult\left(\pi_{k}\right)\\
\nu_{k}\sim Beta(a_{k},b_{k})
\end{array}
\end{equation}

Only conjugate posterior can be learnt the traditional way e.g. MAP
estimate followed by re-arranging a closed form solution. However,
this strategy is not available for non-conjugate case: 

We consider a case of non-conjugate posterior, $\ln q(B{}_{k})$ where
the likelihood is generalized Gaussian distributed and prior Gaussian
distributed. When dealing with conjugate posterior, traditional VI
technique such as the VMM \cite{lim2018fast} take the MAP estimate
to obtain a closed form solution. 

1) Taking the MAP estimate of $\ln q(B_{k})$
\begin{equation}
\begin{array}{c}
E\left[B_{k}\right]=\underset{B_{k}}{\arg\max}\;\ln q(B_{k})\\
=\underset{B_{k}}{\arg\max}\;\underset{z_{nk}}{E}\left[\ln p(x_{n}\mid B_{k},z_{nk})+\ln p(B_{k})\right]
\end{array}
\end{equation}

2) Because the likelihood is not from the exponential family, re-arranging
the gradient in terms of $B_{k}$ for $\nabla_{B_{k}}\ln q(B_{k})=0$
is difficult
\begin{equation}
\begin{array}{c}
\nabla_{B_{k}}\ln p(x_{n}\mid B_{k},E\left[z_{nk}\right])\\
\\
=\frac{\rho}{s}\left|\frac{x_{n}-B_{k}}{s}\right|^{\rho-1}sgn(\frac{x_{n}-B_{k}}{s})E\left[z_{nk}\right]
\end{array}
\end{equation}

The above requires i) a numerical approach and ii) a converging learner
for large sample size and large class number. Both problems are the
main highlights of this work and shall be discussed in detail in the
next section.

\section{Proposed learning: Adaptive Stepsize for Variational Inference }

Previously, the goal in (2) and (3) is to learn $\ln q(\theta_{j})$
by deriving a closed-form expression for $E\left[\theta_{j}\right]$.
Unfortunately, this is impossible unless $\ln q(\theta_{j})$ is a
conjugate posterior. In this section, we propose to estimate non-conjugate
posterior using the stochastic gradient ascent (SGA) approach. We
also seek stochastic learning, faster convergence and returning better
local maxima. For the sake of brevity, we refer to $\theta$ as $\theta_{j}$
in this section. 

\subsection{Constant Stepsize SGA for Variational Inference}

To overcome the lack of a closed-form solution for $E\left[\theta\right]$
in (2), some recent works \cite{ranganath2014black,paisley2012variational,kingma2014stochastic,welling2011bayesian,rezende2015variational,knowles2015stochastic},
propose the learning of non-conjugate posterior using Monte Carlo
gradient estimate, $\nabla_{\theta}E\left[f(\theta)\right]\approx\frac{1}{S}\sum_{s=1}^{S}f(\theta)\nabla_{\theta}\ln q(\theta_{s})$
for approximation. However, this approximation is associated with
large gradient variance and requiring generating posterior samples,
$\theta_{s}$. A more recent work \cite{mandt2017stochastic} proposed
using constant stepsize SGA for VI. Similarly, we can re-express the
expectation of $\ln q(\theta)$ using constant stepsize SGA below
(since approaching the local maximum has the same goal as maximizing
the VLP globally)
\begin{equation}
\begin{array}{c}
E\left[\theta\right]_{t}=\int\theta_{j}\,q(\theta_{j})\,d\theta_{j}\\
=E\left[\theta\right]_{t-1}+\eta\nabla_{\theta}\ln q(\theta)
\end{array}
\end{equation}

For SGA, we refer to the gradient of $\ln q(\theta)$ at iteration
$t$ using a minibatch with sample size $M$ as
\begin{equation}
g_{t}=\frac{1}{M}\sum_{m=1}^{M}\nabla_{\theta}\ln q(\theta_{m})
\end{equation}

\subsection{Adaptive Stepsize SGA for Variational Inference}

In stochastic learning, we draw a small subset of samples (e.g. \textgreater 1K
samples) per iteration to update each posterior. This is more effective
than taking the entire dataset (eg. \textgreater 100K samples) for
learning. In stochastic optimization \cite{robbins1985stochastic},
there is a requirement for a decaying step-size $p_{t}$ to ensure
convergence in SGA as given by $\sum p_{t}=\infty$ and $\sum p_{t}^{2}<\infty$.
This is to avoid SGA bouncing around the optimum of the objective
function. 

In SVI \cite{Hoffman2013}, the main goal is to obtain the ``global
parameter'' update of conjugate posterior from its ``immediate global
parameter'' as $\left(\phi_{global}\right)_{t}=(1-p_{t})\left(\phi_{global}\right)_{t-1}+p_{t}\cdot\phi_{immed}$.
The ``immediate global parameter'' is defined as a noisy estimate
and is cheaper to run since it is computed from a data point sampled
each iteration, rather than from the whole data set. The decaying
step-size is defined as $p_{t}=\left(\tau+t\right)^{-\kappa}$ and
both $\tau$ and $\kappa$ are treated as constants. Our view of the
SVI update equation above is much simpler and has little to do with
SVI. Instead, we simply treat it as a weighted average between current
and previous computed gradient of the posterior to ensure convergence
in learning. In fact, in Table I we observe that SVI has a similar
moment form to the common technique called ``SGA with momentum''.
The main difference being $p_{t}$ is a decaying term rather than
fixed constant e.g. $\beta_{1}$. Thus, we define the \textbf{first
moment of the gradient }(of the posterior) for a given minibatch of
size $M$ samples at $t$ iteration as
\begin{equation}
W_{t}=\left(1-p_{t}\right)W_{t-1}+p_{t}\cdot g_{t}
\end{equation}

The non-conjugate learner in  (4) is based on the SGA approach. Since
we are dealing with an approximate posterior or posterior which is
assumed convex, a more superior gradient learning is the natural gradient
learning. Natural gradient learning is superior to plain vanilla gradient
learning because the shortest path between two point is not a straight-line
but instead falls along the curvature of the posterior objective \cite{amari1998natural}.
Natural stochastic gradient ascent of posterior \cite{honkela2007natural,Hoffman2013}
is defined as $\begin{array}{c}
E\left[\theta\right]_{t}=E\left[\theta\right]_{t-1}+\eta G^{-1}\nabla_{\theta}\ln q(\theta)\end{array}$, where Fisher information matrix $G=E\left[\nabla_{\theta}\ln q(\theta)\left(\,\nabla_{\theta}\ln q(\theta)\,\right){}^{T}\right]$.
The motivation for the steepest ascent direction search of posterior
optimum is best explained by Riemannian geometry in \cite{amari1998natural,Hoffman2013,honkela2007natural}.
When we assumed each dimension is independent (spherical or diagonal),
we end up with the squared gradient of posterior, $G=E\left[\left(\,\nabla_{\theta}\ln q(\theta)\,\right)^{2}\right]$.
For a minibatch of size $M$ samples, we introduce the \textbf{second
moment of the gradient} for the squared gradient of posterior, using
the identity $E\left[X^{2}\right]\geq\left\{ E\left[X\right]\right\} ^{2}$
as follows
\begin{equation}
F_{t}=\left(1-p_{t}\right)F_{t-1}+p_{t}\cdot g_{t}^{2}
\end{equation}

We can take the product of the \textbf{first moment of the gradient}
in (6) and the \textbf{second moment of the gradient} in (7) together
to obtain an \textbf{adaptive} \textbf{stepsize} update 
\begin{equation}
E\left[\theta\right]_{t}=E\left[\theta\right]_{t-1}+\eta\frac{W_{t}}{\sqrt{F_{t}+\epsilon}}
\end{equation}

We defined $\epsilon=10^{-8}$. In the next section, we will make
comparison on (8) with other SGAs.

\subsection{Motivation and Comparison with SGAs}

Our adaptive stepsize learner is motivated by recent SGA methods and
SVI as summarized in Table I. We briefly discuss their similarity
below using the case of $\ln q(\theta)$. 

\uline{SVI:} In Table I, we compare  (6) to the 1st moment in SVI.
We can view the closed-form estimate $\hat{\theta}$ as $g_{t}$ while
$E\left[\theta\right]_{t}$ is seen as $W_{t}$ in (6).

\uline{Momentum SGA:} Similarly, when we fix $p_{t}$ with a constant
value (e.g. at iteration 45 in Fig 1.) over the decaying value in
(6), both $W_{t}$ in (6) and $S_{t}$ will have very similar 1st
moment in Table I.

\uline{Adam:} At a glance, Adam appears to be similar to Momentum
SGA for both their 1st moment. The only difference is that Adam normalize
it with a decaying curve e.g. $\beta_{1}^{t}$. Thus, when we take
an instaneous value in Fig 1, the value of $M_{t}$ is proportional
to $S_{t}$ and vice versa for $W_{t}$. Our definition of $W_{t}$
and $F_{t}$ look very similar to $M_{t}$ and $V_{t}$ in Adam. The
main difference lies in the way we define the stepsizes $p_{t}$.
We adopt the decreasing stepsize defined by SVI. We also use an identical
expression to Adam for the adaptive stepsize update in (8). 

\begin{table*}
\caption{Comparison of learners for variational inference (SVI) and neural
network (SGA)}

\centering{}%
\begin{tabular}{|c|c|c|c|c|}
\hline 
 & Methods & 1st moment of Gradient & 2nd moment of Gradient & Stepsize\tabularnewline
\hline 
\multirow{5}{*}{VI} &  &  &  & \tabularnewline
 & SVI & $E\left[\theta\right]_{t}=\left(1-p_{t}\right)\cdot E\left[\theta\right]_{t-1}+p_{t}\cdot\hat{\theta}$ & - & -\tabularnewline
 &  &  &  & \tabularnewline
\cline{2-5} \cline{3-5} \cline{4-5} \cline{5-5} 
 &  &  &  & \tabularnewline
 & Proposed & $W_{t}=\left(1-p_{t}\right)\cdot W_{t-1}+p_{t}\cdot g_{t}$ & $F_{t}=\left(1-p_{t}\right)\cdot F_{t-1}+p_{t}\cdot g_{t}^{2}$ & $E\left[\theta\right]_{t}=E\left[\theta\right]_{t-1}+\eta\frac{W_{t}}{\sqrt{F_{t}+\epsilon}}$\tabularnewline
 &  &  &  & \tabularnewline
\hline 
\hline 
\multirow{9}{*}{Non-VI} & \multirow{3}{*}{SGA} &  &  & \tabularnewline
 &  & - & - & $E\left[\theta\right]_{t}=E\left[\theta\right]_{t-1}+\eta\frac{g_{t}}{1}$\tabularnewline
 &  &  &  & \tabularnewline
\cline{2-5} \cline{3-5} \cline{4-5} \cline{5-5} 
 & \multirow{3}{*}{Momentum SGA} &  &  & \tabularnewline
 &  & $S_{t}=\beta_{1}\cdot S_{t-1}+\left(1-\beta_{1}\right)\cdot g_{t}$ & - & $E\left[\theta\right]_{t}=E\left[\theta\right]_{t-1}+\eta\frac{S_{t}}{1}$\tabularnewline
 &  &  &  & \tabularnewline
\cline{2-5} \cline{3-5} \cline{4-5} \cline{5-5} 
 &  &  &  & \tabularnewline
 & Adam & $M_{t}=\frac{\beta_{1}\cdot M_{t-1}+\left(1-\beta_{1}\right)\cdot g_{t}}{1-\beta_{1}^{t}}$ & $V_{t}=\frac{\beta_{2}\cdot V_{t-1}+\left(1-\beta_{2}\right)\cdot g_{t}^{2}}{1-\beta_{2}^{t}}$ & $E\left[\theta\right]_{t}=E\left[\theta\right]_{t-1}+\eta\frac{M_{t}}{\sqrt{V_{t}+\epsilon}}$\tabularnewline
 &  &  &  & \tabularnewline
\hline 
\end{tabular}
\end{table*}

\subsection{Brief analysis on convergence}

We plot the curves for $\left(1-p_{t}\right)$ and $p_{t}$ to exhibit
the behavior of using these stepsizes for $W_{t}$ or $F_{t}$. We
set the values $\tau=1$ and $\kappa=0.5$ for $t=50$ iterations
in Fig 1. As the number of iterations increases, for $W_{t}$ and
$F_{t}$, we see that the curves gradually shift responsibilities
from the gradual diminishing value of $p_{t}$ to the increasing value
of $\left(1-p_{t}\right)$.

Recall that in the SVI update $E\left[\theta\right]_{t}=\left(1-p_{t}\right)E\left[\theta\right]_{t-1}+p_{t}\hat{\theta}$,
the term $\hat{\theta}$ is defined as the closed form coordinate
ascent estimate in \cite{Hoffman2013}. Alternatively, $\hat{\theta}$
is computed identical to the conjugate posterior using VMM. Thus,
when we let $\hat{\theta}=\nabla_{\theta}\ln q(\theta)$ at $\nabla_{\theta}\ln q(\theta)=0$
we have the following for SVI
\begin{equation}
E\left[\theta\right]_{t}=\left(1-p_{t}\right)E\left[\theta\right]_{t-1}+p_{t}\nabla_{\theta}\ln q(\theta)
\end{equation}

For the proposed adaptive stepsize in (8), we only discuss the case
of $E\left[\theta\right]_{t}=E\left[\theta\right]_{t-1}+\eta W_{t}$.
Expanding the terms inside, we have the following
\begin{equation}
E\left[\theta\right]_{t}=E\left[\theta\right]_{t-1}+\eta\left(1-p_{t}\right)W_{t-1}+\eta p_{t}\nabla_{\theta}\ln q(\theta)
\end{equation}

Given that $\lim_{t\rightarrow\infty}\left(1-p_{t}\right)=1$ and
$\lim_{t\rightarrow\infty}p_{t}=0$ in Fig 1, we can see that SVI
becomes
\begin{equation}
\lim_{t\rightarrow\infty}E\left[\theta\right]_{t}=E\left[\theta\right]_{t-1}
\end{equation}
while (8) becomes
\begin{equation}
\lim_{t\rightarrow\infty}E\left[\theta\right]_{t}=E\left[\theta\right]_{t-1}+\eta W_{t-1}
\end{equation}

(11) shows that SVI will reach convergence if $E\left[\theta\right]_{t-1}$
is a convex function. (12) consists of an additional term apart from
$E\left[\theta\right]_{t-1}$. Specifically, $W_{t-1}$ consists of
a weighted sum between $\nabla_{\theta}\ln q(\theta)$ and the previous
$W_{t-1}$. Thus, as long as $\nabla_{\theta}\ln q(\theta)$ is a
convex function we can sufficiently ensure that the proposed stepsize
in (8) will also converge. 

\begin{figure}
\begin{centering}
\includegraphics[scale=0.35]{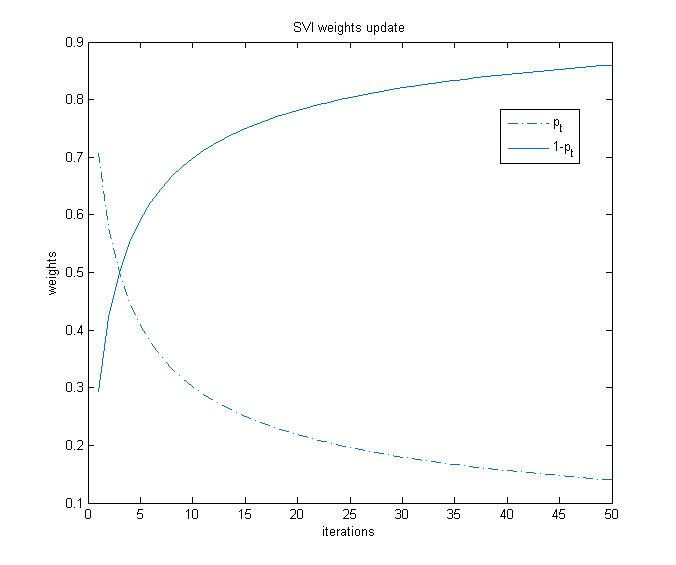}
\par\end{centering}
\caption{Behavior of stepsizes using $p_{t}=\left(1+t\right)^{-0.5}$}
\end{figure}

\section{Proposed Inference of PYPM}

We are ready to perform PYPM inference on a dataset given the expectation
of all three posterior types (non-conjugate, discrete, conjugate)
can be solved. Essentially, we repeat the estimation of all expectations
using minibatch each iteration till convergence or sufficient iterations
has passed. First, we turn to some formalities on PYPM and GGD. Second,
we discuss our proposed inference of PYPM.

\subsection{Pitman-Yor Process }

For the last decade, Dirichlet process Gaussian mixture (DPM) has
mainly found application in model selection of classification datasets
such as UCI, MNIST, text classification, object recognition, scene
recognition and etc. The model selection aspect of DPM actually comes
from Dirichlet process while the distribution of each component of
the mixture comes from a Gaussian. Both Dirichlet process and Gaussian
mixture in DPM are assumed disjointed in VI. Another view of Dirichlet
process is to consider it as a specific case of the Pitman-Yor process
\cite{teh2010hierarchical}. The latter can model additional tail
behavior of dataset over Dirichlet process. The Pitman-Yor process
is controlled by a two parameter Beta distribution where the parameters
are$\begin{array}{c}
a_{k}=1-d\end{array}$ and $b_{k}=\alpha_{0}+kd$ for $0\leq d<1$
\begin{equation}
Beta(v_{k};a_{k},b_{k})\propto v_{k}^{\left(a_{k}-1\right)}\left(1-v_{k}\right)^{\left(b_{k}-1\right)}
\end{equation}
If we set $d=0$ in the above expression then Pitman-Yor process reduces
back to the Dirichlet process. 

\subsection{Generalized Gaussian Density }

In GGD, cluster mean is denoted $B=\left\{ B_{k}\right\} _{k=1}^{K}\in\mathbb{R}{}^{D}$
and we have two new hidden variables, shape and scale. They are $s=\left\{ s_{k}\right\} _{k=1}^{K}\in\mathbb{R}{}^{D}$
and $\rho=\left\{ \rho{}_{k}\right\} _{k=1}^{K}\in\mathbb{R}{}^{D}$
respectively. Specific cases of GGD are the Gaussian PDF $\left(s=\sqrt{2},\rho=2\right)$
and Laplacian PDF $\left(s=\sqrt{2},\rho=1\right)$. Although, the
GGD can be solved by the method of moments for $s$ and $\rho$, there
is no closed-form parameter estimation for GGD when $B$ is non zero
centered. In this work, we are only interested in exploring a new
non-conjugate form to replace GMM. Hence for functionality, we limit
our learning to $B$, while fixing the parameters $s,\rho.$ The GGD
pdf is defined as follows
\begin{equation}
\mathcal{GGD}(x|B,s,\rho)\propto\exp\left(-\left|\frac{x-B}{s}\right|^{\rho}\right)
\end{equation}

\subsection{PYPM}

The joint probability of PYPM can be depicted as $p(x,B,z,v)=p(x\mid B,z)p(B)p(z\mid v)p(v)$.
The observation is denoted $x=\left\{ x_{n}\right\} _{n=1}^{N}\in\mathbb{R}{}^{D}$.
The cluster assignment is denoted $z=\left\{ z_{n}\right\} _{n=1}^{N}$
where $z_{n}$ is a $1-of-K$ binary vector, subjected to $\sum_{k=1}^{K}z_{nk}=1$
and $z_{nk}\in\left\{ 0,1\right\} $. We have earlier summarized the
distribution of each term in PYPM in (1). 

\textbf{Non-Conjugate Posterior:} The stochastic learning of PYPM
is obtained by the proposed sVMM procedure for updating the generalized
Gaussian-Gaussian posterior, $E\left[B_{k}\right]_{t}=E\left[B_{k}\right]_{t-1}+\eta\frac{W_{t}}{\sqrt{F_{t}}}$,
whereby $g_{t}=\frac{1}{M}\sum_{m=1}^{M}\nabla_{B_{k}}\ln q(B_{k})$.
Due to requiring an initial or previous estimate, the non-conjugate
posterior's gradient is computed as follows
\begin{equation}
\begin{array}{c}
\nabla_{B_{k}}\ln q(B_{k})=\frac{\rho}{s}\left|\frac{x_{n}-E\left[B_{k}\right]_{t-1}}{s}\right|^{\rho-1}sgn(\frac{x_{n}-E\left[B_{k}\right]_{t-1}}{s})E\left[z_{nk}\right]\\
-\lambda_{0}(E\left[B_{k}\right]_{t-1}-B_{0})
\end{array}
\end{equation}

\textbf{Discrete Posterior:} In VMM, we update the two conditional
density by running through all possible $K$ states of $z_{n}$ that
maximizes the posterior as below
\begin{equation}
\begin{array}{c}
E\left[z_{nk}\right]=\underset{z_{nk}}{\arg\max}\;\underset{B_{k},v_{k}}{E}\left[\ln p(x_{n}\mid B_{k},z_{nk})+\ln p(z_{nk}\mid v_{k})\right]\\
=\underset{z_{nk}}{\arg\max}\;-\left\{ \left|\frac{x_{n}-E\left[B_{k}\right]}{s}\right|^{\rho}\right.\\
\left.+\ln E\left[v_{k}\right]+\sum_{l=1}^{k-1}\ln(1-E\left[v_{l}\right])\right\} z_{nk}
\end{array}
\end{equation}

\textbf{Conjugate Posterior:} Using VMM, we apply the MAP estiamte
and re-arrange it to obtain a closed form for updating the Multinomial-Beta
posterior below
\begin{equation}
E\left[v_{k}\right]=\frac{\sum_{n=1}^{N}E\left[z_{nk}\right]+\left(a_{k}-1\right)}{\sum_{n=1}^{N}\sum_{j=k+1}^{K}E\left[z_{nj}\right]+\left(a_{k}-1\right)+\left(b_{k}-1\right)}
\end{equation}

We summarized our inference of PYPM in Algo. 1.

\begin{algorithm}
a) Input: $x\leftarrow\left\{ minibatch\right\} $

b) Output: $E\left[z_{nk}\right]$

c) Initialization: $E\left[z_{nk}\right],m_{0},\alpha_{0},\lambda_{0},a_{k},b_{k},K$

d) Repeat update until convergence,
\begin{verse}
1) non-conjugate posterior:
\end{verse}
\[
E\left[B_{k}\right]_{t}=E\left[B_{k}\right]_{t-1}+\eta\frac{W_{t}}{\sqrt{F_{t}}}
\]

\begin{verse}
2) discrete posterior:

\[
\begin{array}{c}
E\left[z_{n}\right]=\underset{z_{nk}}{\arg\max}\;\ln q(z_{n})\end{array}
\]

3) conjugate posterior:

\[
E\left[v_{k}\right]\approx\hat{v_{k}}
\]
\end{verse}
\caption{Proposed Inference of PYPM}
\end{algorithm}

\section{Experiments }

\textbf{Proposed Variants: }We consider three variants of proposed
method in Table IV-VIII as shown below. 
\begin{verse}
1) (Gau: SGA) SGA using (4) for solving $E\left[B_{k}\right]_{t}$,
with Gaussian case where $s=\sqrt{2},\rho=2$ in $\mathcal{GGD}(B,s,\rho)$ 

2) (Gau: AdaSGA) Using our adaptive stepsize in (8) i.e. $E\left[B_{k}\right]_{t}=E\left[B_{k}\right]_{t-1}+\eta\frac{W_{t}}{\sqrt{F_{t}}}$
with Gaussian case as in variant 1. 

3) (Lapl: AdaSGA) similar to variant 2 but now repeated with Laplacian
case where $s=\sqrt{2},\rho=1$ in $\mathcal{GGD}(B,s,\rho)$
\end{verse}
\textbf{Strong Baseline:} We implemented a strong baseline ``SVI:
DPM'' to compare with our best proposed method. This baseline is
the Dirichlet process Gaussian mixture and is also classified under
BNP. It is implemented using the SVI update in Table I, after obtaining
the closed-form expectation of posterior as found in \cite{lim2018fast}.
The remainder of the DPM algorithm is identical to the proposed DPM
algorithm in \cite{lim2018fast}, but without the precision posterior.
We ran at least 10 reruns and took their average (the values inside
the bracket is their standard deviation).

\textbf{Feature:} DDPM-L and OnHGD are using the 128 dimensional SIFT
features. For LDPO, the authors use 4096 dimensional AlexNet pretrained
on ImageNet. DAEC, DC-Kmeans, DC-GMM and DEC are end-to-end models
that rely on the pre-trained and fine-tuned encoder to perform feature
extraction. In comparison, we use the 512 dimensional ResNet18 pretrained
on ImageNet. 

\textbf{Truncation:} For LDPO, DAEC, DC-Kmeans, DC-GMM, DEC, DBC,
it is fixed to the ground truth. For SVI: DPM the truncation setting
are identical to this work. Ground truth refers to the number of classes
per dataset. It ranges from 15 to 397 classes (or clusters in our
case). For unsupervised learning we do not require class labels for
learning our models. However, we require setting a truncation level
(upper limit) for each dataset as our model cannot start with an infinite
number of clusters in practice. We typically use a very large truncation
value (e.g. $K=1000$ for SUN397) away from the ground truth to demonstrate
that our model is not dependent on ground truth information.

\textbf{Datasets: }The datasets used in our experiments are detailed
in Table II. There are 3 scene and 2 digit classification datasets
in total. The largest dataset has about over 100K images, smallest
dataset is at over 4K. We split the datasets into train and test partition. 

\textbf{Minibatch:} For calculating our minibatch size, we approximate
it by $M=sampleperclass*(gnd.truth)$, where $sampleperclass$ is
typically 20 or 30 (for the datasets in this work) for sufficient
statistics. In order to make the training dataset unbiased, we further
assume each set of minibatch has sufficient sample draw from each
class. This is necessary as some dataset have classes with 8000 samples
while other classes have only 100 samples. 

\textbf{Evaluation Metric:} We compare three criteria: i) Normalized
Mutual Information, iii) Accuracy and iv) Model Selection. We use
Normalized Mutual Information (NMI) and Accuracy (ACC) to evaluate
the learning performance of our model. Model refers to the model selection
estimated by each approach. The definition for ACC and NMI are $ACC=\frac{\sum_{n=1}^{N}\delta\left(gt_{n},\:map\left(mo_{n}\right)\right)}{N}$
and $NMI=\frac{MU_{info}(gt,mo)}{\max\left(H\left(gt\right),H\left(mo\right)\right)}$
where $gt,mo,map,\delta\left(\cdotp\right),MU_{info},H$ refers to
ground truth label, model's predicted label, permutation mapping function,
delta function, mutual information and entropy respectively. Delta
function is defined as $\delta(gt,mo)=1$ if $gt=mo$ and equal 0
otherwise. 

\begin{table}
\begin{centering}
\begin{tabular}{|c|c|c|c|c|c|c|}
\hline 
\# & Dataset & Classes & Train & Test & Trunc.  & Minibatch \tabularnewline
 &  &  &  &  & Level & Size\tabularnewline
\hline 
\hline 
1 & Scene15 & 15 & 750 & 3735 & 50 & 300\tabularnewline
\hline 
2 & MIT67 & 37 & 3350 & 12,270 & 100 & 1340\tabularnewline
\hline 
3 & SUN397 & 397 & 39,700 & 69,054 & 1000 & 11,910\tabularnewline
\hline 
4 & MNIST & 10 & 60,000 & 10,000 & 50 & 200\tabularnewline
\hline 
5 & USPS & 10 & 7,291 & 2,007 & 50 & 200\tabularnewline
\hline 
\end{tabular}
\par\end{centering}
$\;$

\caption{Datasets (scene) for Bayesian nonparametrics}

$\;$

$\:$
\begin{centering}
\begin{tabular}{|c|c|c|c|c|}
\hline 
\# & Methods & Year & Feature & Minibatch\tabularnewline
\hline 
\hline 
1 & Kmeans \cite{wang2017unsupervised} & 2017 & AlexNet & no\tabularnewline
\hline 
2 & LDPO-A-FC \cite{wang2017unsupervised} & 2017 & AlexNet & no\tabularnewline
\hline 
3 & OnHGD \cite{fan2016online} & 2016 & SIFT & yes\tabularnewline
\hline 
4 & SVI: DPM \cite{Hoffman2013} & 2013 & ResNet & yes\tabularnewline
\hline 
5 & DAEC \cite{song2013auto} & 2013 & End-to-end & yes\tabularnewline
\hline 
6 & DC-Kmeans \cite{tian2017deepcluster} & 2017 & End-to-end & yes\tabularnewline
\hline 
7 & DC-GMM \cite{tian2017deepcluster} & 2017 & End-to-end & yes\tabularnewline
\hline 
8 & DEC \cite{xie2016unsupervised} & 2016 & End-to-end & yes\tabularnewline
\hline 
9 & DBC \cite{li2018discriminatively} & 2018 & End-to-end & yes\tabularnewline
\hline 
10 & ClusterGAN \cite{ghasedi2019balanced} & 2019 & End-to-end & yes\tabularnewline
\hline 
11 & DASC \cite{zhou2018deep} & 2018 & End-to-end & yes\tabularnewline
\hline 
\end{tabular}
\par\end{centering}
$\;$

\caption{Recently published methods used in this comparison}
\end{table}

\subsection{Comparison with Bayesian Nonparametrics }

\textbf{Bayesian nonparametrics (BNP):} BNPs can perform clustering
and estimate the cluster number jointly. The work here solely consider
the pursuit of advancing statistical model for large scale datasets.
The method are OnHGD (based on SVI) \cite{fan2016online} and our
baseline method SVI: DPM. Our work is also categorized under this
area. 

We compare our work with recent works citing the datasets we use.
First, we group the published methods using the dataset in Table III.
Next, we compare some of these published methods (non end-to-end)
with our proposed variants in Table IV-VI. Also, we use 10 reruns
for our proposed method and took their average. We can achieve convergence
on our proposed variants with around 100 iterations. 

\subsubsection{Scene15}

In Table IV, Gau: SGA is able to outperform LDPO-A-FC and SVI: DPM.
and Kmeans. When adaptive stepsize is applied in Gau: AdaSGA, it further
improves the clustering and model selection result. Lapl: AdaSGA is
unable to significantly outperform Gau: AdaSGA for this dataset as
the sample size is quite small. 

\subsubsection{MIT67}

To the best of our knowledge, it is very rare to find recent deep
clustering works (e.g. \cite{tian2017deepcluster,caron2018deep})
addressing datasets beyond 10 classes for image datasets. The main
reason we suspect is that most recent related works rely on end-to-end
learning (i.e. the encoder of the autoencoder) rather than use an
ImageNet pretrained CNN for feature extraction. It is likely more
difficult to train or finetune the encoder to be as discriminative
as ResNet especially when there is only about 200 samples per class
for MIT67 in Table III.

In Table V, LDPO-A-FC is almost on par with its baseline comparison
using Kmeans on the larger MIT67 dataset at ACC of 37.9\% vs 35.6\%
respectively. Our baseline method ``SVI: DPM'' using ResNet18 feature
also perform better than LDPO-A-FC at ACC of 61.21\%. We outperformed
the best published method by almost double in performance using ``Lapl:
AdaSGA'' at ACC of 64.47\%. We also notice that SVI: DPM outperforms
Gau: SGA and Gau: AdaSGA. We believe SVI: DPM works better on larger
dataset and the benefit of using a closed form solution is definitely
more robust than a numerical approach such as SGA with all things
being equal. Fortunately, Lapl: AdaSGA turns the verdict around by
offering a more discriminative model that surpasses Gaussian for this
dataset. The stronger model and the adaptive stepsize both attribute
to the best performance of Lapl: AdaSGA on MIT67.

\subsubsection{SUN397}

In Table VI, OnHGD applies SVI \cite{Hoffman2013} (``On'' for online)
to their BNP model HGD. They use OnHGD to learn a Bag-of-Words representation
for SUN397. It appears they then use a supervised learner such as
Bayes's decision rule for classification. No model selection was mentioned
for SUN397 either. For SUN397, the ACC reported in OnHGD was 26.52\%
on SUN397. Although this is not a direct comparison, the same authors
also reported an ACC of 67.34\% for SUN16. In comparison, we obtained
83.37\% on Scene15.

Our baseline ``SVI: DPM'' was able to get 39.07\% on ACC compared
to OnHGD of 26.52\%. Both Gau: SGA and Gau: AdaSGA are performing
worse that SVI: DPM. This is another evidence that SGA is inferior
to SVI. Adaptive stepsize can help reduce the gap. Our best result
is ``Lapl: AdaSGA'' which was able to slightly improve the results
to 40.39\% on the same dataset. The saving grace most likely being
the discriminative power of the Laplacian mixture model. 

Although not shown, the convergence of ``Lapl: AdaSGA'' is much
slower for this particular dataset. Due to computational budget, we
did not further check if better ACC can be obtained beyond 200 iterations
using Algo 1. Also, our implementation for ``SVI: DPM'' faced some
cluster singularity issue (cluster disappearing) when given too many
iterations for SUN397. We had to stop iterations after around 15 or
20 as the cluster count may fall below 397.

\begin{table}
\begin{centering}
\begin{tabular}{|c|c|c|c|}
\hline 
 & NMI & ACC & Model\tabularnewline
\hline 
\hline 
Kmeans \cite{wang2017unsupervised} & 0.659 & 0.65 & -\tabularnewline
\hline 
LDPO-A-FC \cite{wang2017unsupervised} & 0.705 & 0.731 & -\tabularnewline
\hline 
SVI: DPM (baseline) & 0.7877 & 0.7659 & -\tabularnewline
\hline 
Gau: SGA (ours) & 0.80333 & 0.81901 & 22\tabularnewline
\hline 
Gau: AdaSGA (ours) & 0.81201 & \textbf{0.83614} & 21\tabularnewline
\hline 
Lapl: AdaSGA (ours) & \textbf{0.8165} & 0.8337 & \textbf{21}\tabularnewline
\hline 
\end{tabular}
\par\end{centering}
$\;$

\caption{Performance on Scene15}

$\;$

$\;$
\begin{centering}
\begin{tabular}{|c|c|c|c|}
\hline 
 & NMI & ACC & Model\tabularnewline
\hline 
\hline 
Kmeans \cite{wang2017unsupervised} & 0.386 & 0.356 & -\tabularnewline
\hline 
LDPO-A-FC \cite{wang2017unsupervised} & 0.389 & 0.379 & -\tabularnewline
\hline 
SVI: DPM (baseline) & 0.6858 & 0.6121 & -\tabularnewline
\hline 
Gau: SGA (ours) & 0.66106 & 0.56496 & 78\tabularnewline
\hline 
Gau: AdaSGA (ours) & 0.68546 & 0.60244 & 78\tabularnewline
\hline 
Lapl: AdaSGA (ours) & \textbf{0.7081} & \textbf{0.6447} & \textbf{78}\tabularnewline
\hline 
\end{tabular}
\par\end{centering}
$\;$

\caption{Performance on MIT67}

$\;$

$\;$
\begin{centering}
\begin{tabular}{|c|c|c|c|}
\hline 
 & NMI & ACC & Model\tabularnewline
\hline 
\hline 
OnHGD \cite{fan2016online} & - & 0.2652 & -\tabularnewline
\hline 
SVI: DPM (baseline) & 0.596 & 0.3907 & -\tabularnewline
\hline 
Gau: SGA (ours) & 0.5281 & 0.2612 & 489\tabularnewline
\hline 
Gau: AdaSGA (ours) & 0.58226 & 0.34798 & 513\tabularnewline
\hline 
Lapl: AdaSGA (ours) & \textbf{0.6022} & \textbf{0.4039} & \textbf{487}\tabularnewline
\hline 
\end{tabular}
\par\end{centering}
$\;$

\caption{Performance on SUN397}
\end{table}

\subsection{Comparison with Deep Clustering }

\textbf{Deep clustering}: A hybrid between neural network and statistical
clustering, these works perform clustering in the feature space of
the neural network, most of the works using autoencoder or GAN. These
methods are DAEC \cite{song2013auto}, DC-Kmeans \cite{tian2017deepcluster},
DC-GMM \cite{tian2017deepcluster}, DEC \cite{xie2016unsupervised},
DBC \cite{li2018discriminatively} LDPO \cite{wang2017unsupervised},
DASC \cite{zhou2018deep} and ClusterGAN \cite{ghasedi2019balanced}.
Furthermore in these works, the clustering information further optimize
the weights update in the hidden layers. However, the statistical
clustering employed here are typically the fundamentals ones such
as Kmeans or GMM.

\subsubsection{MNIST}

Most recent end-to-end clustering algorithms focus on digit recognition
(i.e. MNIST and USPS) for experiments. Compared to MIT67 and SUN397,
MNIST is a much easier dataset since the number of classes is mediocre
(10 classes) and there is a large number of training images at 60k.

In Table VII, all the end-to-end methods (DAEC, DC-Kmeans/GMM, DEC,
DBC) train a deep encoder (e.g. x-500-500-2000-10) as feature extractor.
In comparison, we use ResNet feature directly as input to ``SVI:
DPM'' and ``Lapl: AdaSGA''. Table VII shows the comparison between
the published methods and ours on MNIST. For our best approach, ``Lapl:
AdaSGA'', we are able to outperform our strong baseline ``SVI: DPM''
as well as obtain comparable ACC and NMI to the best published result
by DBC or ClusterGAN.

\subsubsection{USPS}

In Table VIII, all the end-to-end methods (DAEC, DC-Kmeans, DEC, DBC)
similarly trains a deep encoder as feature extractor. In comparison,
K-means \cite{tian2017deepcluster} using raw image pixel obtains
45.85\% on ACC. For this particular dataset, we only use raw image
pixel as direct input to both to ``SVI: DPM'' and ``Gau: AdaSGA''.
Our best result using ``Gau: AdaSGA'' consistently outperformed
all published result and strong baseline again on USPS at 80.10\%
on ACC. Our baseline is close behind at 77.63\% on ACC. The best published
method DBC obtained 74.3\% but it outperforms our NMI measure. We
believe the reason why most end-to-end methods cannot perform better
than our methods on USPS even though they are using deep encoder features
while we use pixel intensity is partly due to the comparatively small
training size at 7K compared to say 60K on MNIST. 

\begin{table}
\begin{centering}
\begin{tabular}{|c|c|c|}
\hline 
Methods & NMI & ACC\tabularnewline
\hline 
\hline 
DAEC \cite{song2013auto} & 0.6615 & 0.734\tabularnewline
\hline 
DC-Kmeans \cite{tian2017deepcluster} & 0.7448 & 0.7448\tabularnewline
\hline 
DC-GMM \cite{tian2017deepcluster} & 0.8318 & 0.8555\tabularnewline
\hline 
DEC \cite{xie2016unsupervised} & 0.8273 & 0.8496\tabularnewline
\hline 
DBC \cite{li2018discriminatively} & 0.917 & 0.964\tabularnewline
\hline 
ClusterGAN \cite{ghasedi2019balanced} & 0.890 & 0.950\tabularnewline
\hline 
DASC \cite{zhou2018deep} & 0.780 & 0.804\tabularnewline
\hline 
SVI: DPM (baseline) & 0.9233 & 0.9348\tabularnewline
\hline 
Lapl: AdaSGA (ours) & \textbf{0.9517} & \textbf{0.9580}\tabularnewline
\hline 
\end{tabular}
\par\end{centering}
$\;$

\caption{Comparison on MNIST}

$\;$

$\;$
\begin{centering}
\begin{tabular}{|c|c|c|}
\hline 
Methods & NMI & ACC\tabularnewline
\hline 
\hline 
K-means \cite{tian2017deepcluster} & 0.4503 & 0.4585\tabularnewline
\hline 
DAEC \cite{song2013auto} & 0.5449 & 0.6111\tabularnewline
\hline 
DC-Kmeans \cite{tian2017deepcluster} & 0.5737 & 0.6442\tabularnewline
\hline 
DEC \cite{xie2016unsupervised} & 0.651 & 0.6246\tabularnewline
\hline 
DBC \cite{li2018discriminatively} & 0.724 & 0.743\tabularnewline
\hline 
SVI: DPM (baseline) & 0.6223 & 0.7763\tabularnewline
\hline 
Gau: AdaSGA (ours) & \textbf{0.6507} & \textbf{0.8010}\tabularnewline
\hline 
\end{tabular}
\par\end{centering}
$\;$

\caption{Comparison on USPS}
\end{table}

\section{Conclusion}

The stochastic optimization of VI can be broadly categorized under
two types. The first approach formulates the learning of the posterior
using SGA while the second approach rely on traditional closed-form
learning. In literature, the first approach require generating Monte
Carlo sample from the variational posteriors, which is not practical
for large datasets such as SUN397. The second approach suffers from
the constraint of requiring analytical solution for the variational
posterior expectation but has reported the capability to scale up
to 3.8M samples and 200 classes. In this paper, we target up to about
100K samples and 400 classes using ResNet feature pretrained on ImageNet.
We try to improve on the problems faced in both approaches. We first
began with the constant stepsize SGA approach and in order to make
it computationally efficient, we further stochastic optimization for
VI. Stochastic optimization rely on decreasing step-size for guaranteed
convergence. Inspired by Adam, we explored using first and second
order moments of the gradient so as to achieve a faster convergence.
We test our new stochastic learner on the Pitman-Yor process generalized
Gaussian mixture which does not have closed-form learning for the
posterior for specific case of Laplacian and Gaussian. We showed the
significant performance gained in terms of NMI, ACC, model selection
on large class number datasets such as the MIT67 and SUN397 and on
MNIST and USPS with recent end-to-end deep learning related works.

\bibliographystyle{IEEEtran}
\addcontentsline{toc}{section}{\refname}\bibliography{allmyref}

\end{document}